\documentclass{article}
\usepackage{spconf,amsmath,graphicx}
\usepackage{xcolor, pgfplots, makecell}
\pgfplotsset{xtick style={draw=none}}
\pgfplotsset{compat=1.14}

\title{Detecting Forged Facial Videos using convolutional neural networks}
\name{Neilesh Sambhu and Shaun Canavan}
\address{University of South Florida}

\begin{document}
\maketitle

\begin{abstract}
In this paper, we propose to detect forged videos, of faces, in online videos. To facilitate this detection, we propose to use smaller (fewer parameters to learn) convolutional neural networks (CNN), for a data-driven approach to forged video detection. To validate our approach, we investigate the FaceForensics public dataset detailing both frame-based and video-based results. The proposed method is shown to outperform current state of the art. We also perform an ablation study, analyzing the impact of batch size, number of filters, and number of network layers on the accuracy of detecting forged videos.
\end{abstract}


\section{Introduction}
In recent years there has been tremendous progress in manipulating videos, which includes real-time generation, use of audio to synthesize videos, and animating static images. This can undermine applications of biometrics, affective computing, and forms of digital communication such as social media videos and teleconferences. Considering this, we propose a solution to detect forged (i.e. fake) facial videos, using convolutional neural networks (CNN) that are much smaller (fewer parameters to learn) than the current state-of-the-art solutions. The use of smaller networks has the advantage of having less parameters while still being able to learn complex functions similar to deeper networks \cite{ba2014deep}.

Detecting fake facial videos can broadly be categorized into 3 categories (physical, signal, and data-driven). Physical approaches tend to focus on features of the face such as eye blinking and head pose. Li et al. \cite{li2018ictu} used a combination of CNNs and long-term recurrent CNNs to analyze eye blinks. They show that blinking is not well represented in the fake videos, which their proposed network takes advantage of for detection. Yang et al. \cite{yang2019exposing} used a Support Vector Machine (SVM) along with 3D head pose to detect fake face videos. They found that the landmark locations between the original and fake videos differ, due to the synthesis methods used to create the fake videos. For synthesizing videos, Suwajanakorn et al. \cite{suwajanakorn2017synthesizing} investigated synthesizing video from audio around the mouth region. They note that there can be inconsistencies between the lips and speech as they may not be synchronized, resulting in an area of the face for analysis of fake videos (i.e. the mouth can be used, similar to eye blinking). Agarwal et al. \cite{agarwal2019protecting} captured behavioral patterns, from 2D facial landmarks, of individuals to detect Deep Fake videos. From these landmarks, they investigate the occurrence and intensity of Facial Action Units \cite{FACS}, training an SVM, with this data, to detect fake video sequences.

Signal-based approaches tend to focus on artifacts that are introduced during the Deep Fake synthesis (creation) phase. Matern et al. \cite{matern2019exploiting} showed that visual artifacts such as changes in eye color can be reliably exploited to detect fake videos. They characterize the differences in eye color based on the HSV color space, which is used along with a bagged version of k-nearest neighbors, to detect fake face videos. Li et al. \cite{li2018exposing}, developed an approach motivated by the idea that Deep Fake generation algorithms have a limited resolution and require warping. They showed that CNNs can capture this information to distinguish between fake and real videos. 

Data-driven approaches are generally simpler in that they don’t look for specific artifacts but focus on large amounts of training data that contains both real and Deep Fake data. Guera et al. \cite{guera2018deepfake} used convolutional neural networks (CNNs) to extract frame-level features which are then used to train a recurrent neural network to detect whether a video has been manipulated or not. They show that their proposed approach can detect fake vs. real videos with less than 2 seconds of data. Nguyen et al.  \cite{nguyen2019capsule} used capsule networks \cite{sabour2017dynamic} to detect forged videos that include replay attacks \cite{patel2015live}, as well as computer generated (e.g. Generative Adversarial Networks \cite{goodfellow2014generative}). They showed that by adding random Gaussian noise to their network, then can improve the detection accuracy.

The proposed approach to detecting forged facial videos can be categorized as data-driven and is motivated by the works detailed here. Our main contributions are 3-fold and can be summarized as follows:
\begin{enumerate}
    \item A CNN that has fewer parameters to learn is proposed to detect fake face videos. We report frame- and video-level results.
    \item Proposed network outperforms current state of the art on the FaceForensics \cite{rossler2018faceforensics} dataset.
    \item Details on the impact of batch size, number of filters, and number of network layers (i.e. ablation study) on the accuracy of detecting fake face videos are given.
\end{enumerate}
\section{Experimental Design}
\begin{figure}
\includegraphics[width=8.5cm, height=4.5cm]{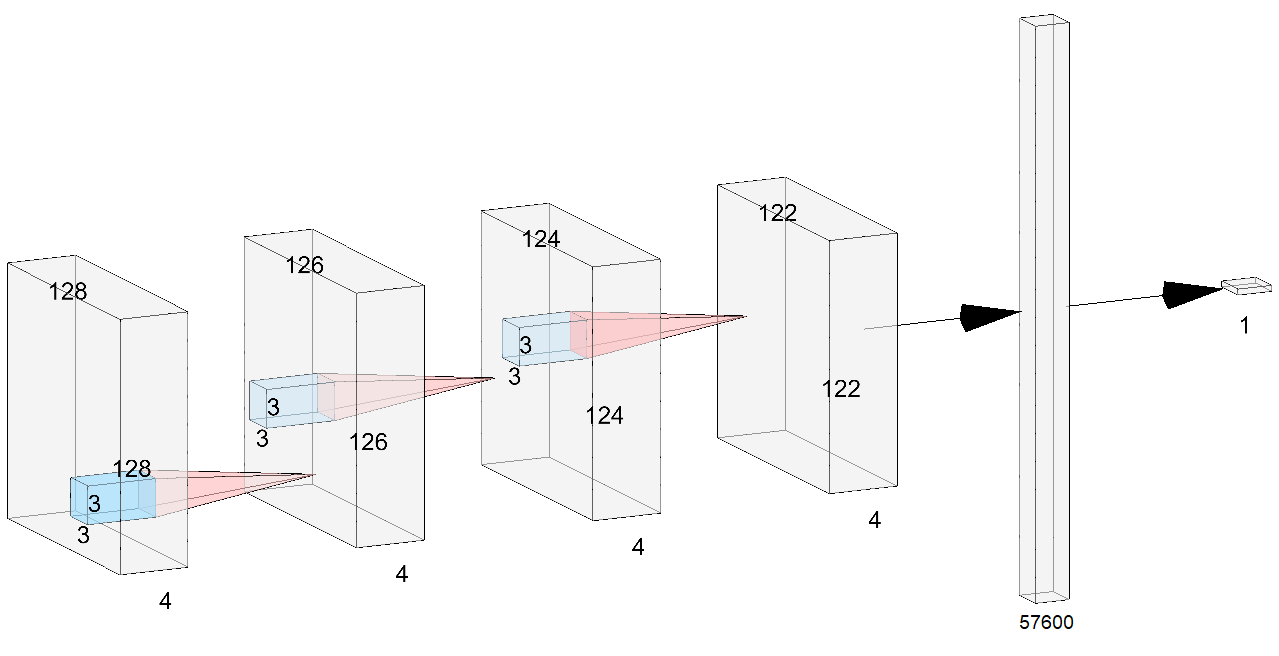}
\caption{Overview of proposed CNN architecture.}
\label{fig:network}
\end{figure}

To detect fake face videos, we propose to use a convolutional neural network (CNN) that is smaller (i.e. less parameters to learn) compared to other state-of-the-art works for detecting fake face videos \cite{rossler2018faceforensics}. To evaluate our proposed CNN, we conduct experiments on the publicly available FaceForensics datasets \cite{rossler2018faceforensics}. Details on our proposed network architecture and this dataset are given in the following subsections.

\subsection{Convolutional Neural Network Architecture}
\label{sec:CNN}
Our proposed CNN, has $5$ layers with $58,221$ parameters. This is compared to other works \cite{rossler2018faceforensics} that use XCeptionNet \cite{chollet2017xception}, which has $71$ layers and approximately $23$ million parameters. In our network, the first 4 blocks represent 2D convolutions, with $4$ filters of size $3\times3$ for each convolutional layer. Batch normalization follows each convolutional layer, with a final dense layer of size $1$. The Adam optimizer \cite{kingma2014adam} with a learning rate of 0.001 was used along with binary cross-entropy as the loss function, and accuracy as the evaluation metric. A batch size of 128 was used and the network was trained for $10$ epochs. We implemented early stopping when the difference in validation accuracy was less than 0.01. As the network outputs a probability between $[0,1]$, we implement a threshold where any value $< 0.5$ is classified as original and any value $\geq 0.5$ is classified as fake. In Section \ref{sec:results}, we detail the impact of batch size, number of filters and layers, on the proposed architecture's ability to detect fake face videos. See Figure \ref{fig:network} for an overview of the proposed architecture.

\subsection{FaceForensics Dataset}
\label{sec:dataset}
\begin{figure}
\includegraphics[width=8.5cm, height=4cm]{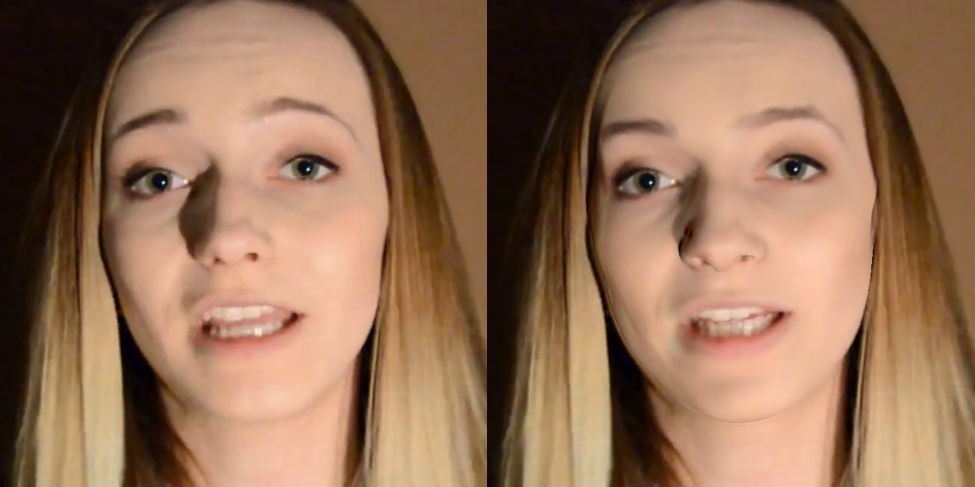}
\caption{Sample images from the FaceForensics dataset \cite{rossler2018faceforensics}. Left: original, right: fake.}
\label{fig:faceForensics}
\end{figure}

The FaceForensics dataset \cite{rossler2018faceforensics} consists of 1004 videos that have a resolution greater than 480p from the youtube8m \cite{abu2016youtube} dataset. Videos that were tagged with labels such as "face" were selected for the dataset. Face2Face \cite{thies2016face2face} was used to create the face forgery videos (called “altered” videos) from the parent videos (called “original” videos). As such, there is perfect class balance in the dataset between altered and original videos. The FaceForensics dataset  is split into training, validation, and testing sets that includes 736,270 samples of training data sourced from 704 videos, 151,052 samples of validation data sourced from 150 videos, and 155,490 samples of testing data sourced from 150 videos. The aforementioned frame count values are split evenly between altered and original videos resulting in a balanced dataset of original and forged videos. For our experiments, we used all of the data in the training, validation, and testing sets. See Figure \ref{fig:faceForensics} for examples of an original and fake image from this dataset.

\section{Results}
\label{sec:results}
To evaluate the utility of the proposed CNN to detect fake videos, we conducted frame- and video-based experiments. We also evaluated the impact of batch size, the number of filters, and the number of convolutional layers.

\subsection{Frame-based Results}
\label{sec:framebased}
\begin{table}
  \centering
  \caption{Confusion matrix for frame-based results on entire testing set from the FaceForensics dataset \cite{rossler2018faceforensics}.}
    \setlength{\tabcolsep}{3pt}
    \begin{tabular}{|c|c|c|}
    \hline
    &Detected Original & Detected Fake \\
    \hline
    Ground Truth Original & \bfseries .999 & .001\\
    \hline
    Ground Truth Fake & .007 & \bfseries .993 \\
    \hline
    \end{tabular}
    \label{table:confMat}
\end{table}
As noted in Section \ref{sec:dataset}, the FaceForensics dataset contains pre-sorted training, validation and testing sets. To conduct our experiments, we trained our proposed network (Section \ref{sec:CNN}) on the entire training set, and here we report our frame-based results (i.e. individual detection result for each video frame) on the entire testing set ($155,490$ images). Using the proposed CNN, we achieve an accuracy of $99.6\%$, where $627$ images were misclassified. As can be seen in Table \ref{table:confMat}, a small percentage of frames where misclassified as fake, when they were original (70 frames). While more frames were misclassified as original, when they were fake (557 frames), the overall accuracy is high for both, showing the proposed smaller CNN is robust to detect fake face videos.

\subsection{Video-based Results}
\label{sec:videoBased}
\begin{figure}
\includegraphics[width=8.5cm, height=4cm]{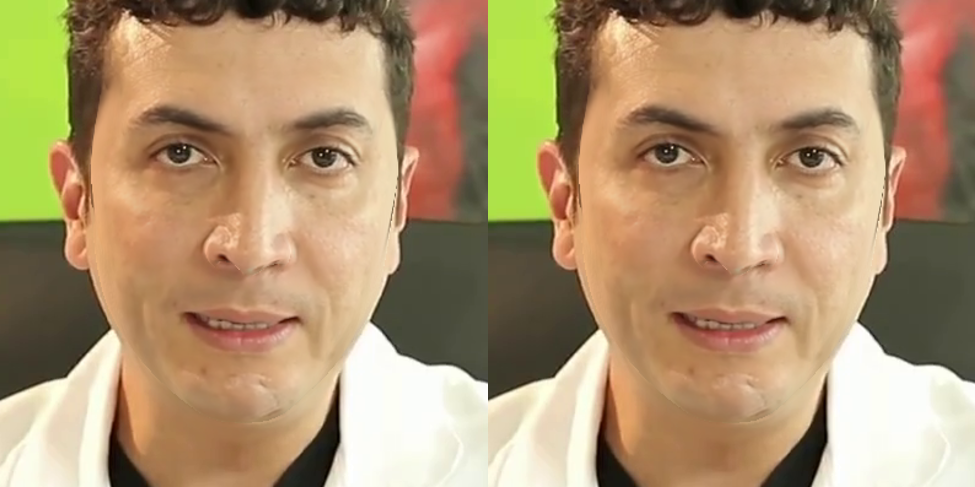}
\caption{Consecutive frames from the 1 misclassified video. Ground truth of video is fake. Left: misclassified as original, right: correctly classified as fake.}
\label{fig:badVideoFrames}
\end{figure}

\begin{table}
  \centering
  \caption{Confusion matrix for video-based results on entire testing set from the FaceForensics dataset \cite{rossler2018faceforensics}.}
    \setlength{\tabcolsep}{3pt}
    \begin{tabular}{|c|c|c|}
    \hline
    &Detected Original & Detected Fake \\
    \hline
    Ground Truth Original & \bfseries 1 & 0\\
    \hline
    Ground Truth Fake & .007 & \bfseries .993 \\
    \hline
    \end{tabular}
    \label{table:confMatVideo}
\end{table}

To conduct our video-based experiments (i.e. detect whether a video is fake or not), we followed the same experimental design as our frame-based detection, however, we also implemented majority voting for each video. For each video, the final classification (real or fake) is calculated by summing up the total classifications for each frame. The classification with the majority of frames labeled as such, is determined to be the final classification. Using majority voting results in a video-based accuracy of $99.67\%$, on the 150 testing videos, with only 1 video being misclassified. As can be seen in Table \ref{table:confMatVideo}, 1 video was detected as original, when it has a ground truth label of fake. As we used majority voting for our video-based detection, this video was incorrectly classified as $53\%$ of the frames were classified as original. 

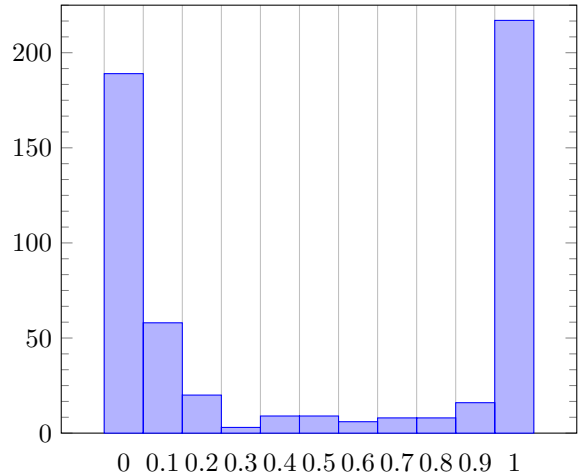
\begin{figure}
\begin{tikzpicture}
\begin{axis}[ybar interval, ymax=225,ymin=0, minor y tick num = 5]
\addplot+[mark=no] coordinates { (0, 189) (0.1, 58) (0.2, 20) (0.3, 3) (0.4, 9) (0.5, 9) (0.6, 6) (0.7, 8) (0.8, 8) (0.9, 16) (1, 217) (1.1, 0)};
\end{axis}
\end{tikzpicture}
\caption{Histogram of probabilities from misclassified video.}
\label{fig:histo}
\end{figure}

To gain further insight into why this video was misclassified, we computed a histogram of the probabilities for each frame in this video. As can be seen in Figure \ref{fig:histo}, the network was largely confident in it's predictions with 189 having a probability of 0 (original), and 217 frames having a probability of 1 (fake). The misclassification occurred, in part, due to the probabilities that were not 0 or 1, where a total of 90 frames were also classified as original, however, with a lower probability. It is also interesting to note that this video contributed to $45\%$ of the misclassified frames from Section \ref{sec:framebased} ($270$ out of $627$ total misclassified frames). Figure \ref{fig:badVideoFrames} shows two consecutive frames from this misclassified video. The frame on the left was misclassified as original, while the frame on the right was correctly classified as fake. As can be seen in this figure, visually the frames look similar, however, the network had different classification labels for each. 

\subsection{Impact of Fine-tuning CNN Architecture}
To evaluate the robustness of the proposed network, we conducted experiments to investigate the impact of batch size, number of filters and number of convolutional layers.

\begin{figure}
\begin{tikzpicture}
	\begin{axis}[
	    ymin=96,
	    ymax=100,
	    xtick={1, 2, 3, 4},
		xlabel=Number of Convolutional Layers,
		ylabel=Accuracy]
	\addplot[mark=*] coordinates {
		(1, 96.9)
		(2, 98.1)
		(3, 99.2)
		(4, 99.3)
	};
	\end{axis}
\end{tikzpicture}
\caption{Impact of number of convolutional layers, on accuracy, in the proposed network.}
\label{fig:ablation}
\end{figure}
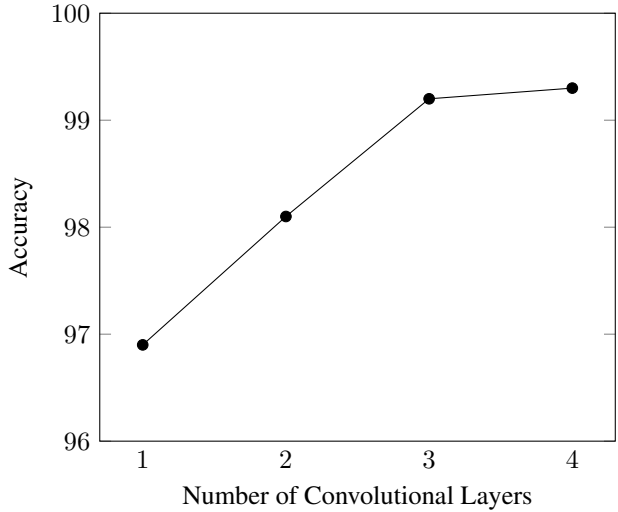

\textbf{Convolutional Layers.} We evaluated the proposed network using 1, 2, 3, and 4 convolutional layers using 4 filters of size $3 \times 3$, with a batch size of 64. The number of layers had little impact on the overall accuracy. As can be seen in Figure \ref{fig:ablation}, increasing the number of layers from 1 to 4 shows an increase for each layer added, however, this increase is $<3\%$ when comparing the lowest accuracy of $96.9\%$ with 1-layer to the max accuracy of $99.3\%$ with 4-layers. These results are encouraging, showing the robustness of smaller CNNs to detect fake face videos. 

\begin{table}
  \centering
  \caption{Impact of batch size on accuracy when training proposed network for 1 epoch.}
    \setlength{\tabcolsep}{3pt}
    \begin{tabular}{|c|c|c|c|}
    \hline
    Batch Size & \makecell{Training\\Accuracy} & \makecell{Validation\\ Accuracy} & \makecell{Testing\\Accuracy} \\
    \hline
    \bfseries 64 & \bfseries \bfseries 0.8394 & \bfseries 0.8137 & \bfseries 0.8051 \\
    \hline
    \bfseries 128 & \bfseries \bfseries 0.8463 & \bfseries 0.8145 & \bfseries 0.8009 \\
    \hline
    256 & 0.8176 & 0.5450 & 0.5627 \\
    \hline
    512 & 0.7525 & 0.6870 & 0.6987 \\
    \hline
    1024 & 0.6843 & 0.6096 & 0.5796 \\
    \hline
    \end{tabular}
    \label{table:batch}
\end{table}
\textbf{Batch Size.} Batch size in a CNN can impact the time to converge, as well as overfitting of the network \cite{radiuk2017impact}. Considering this, we evaluated the impact of batch size on accuracy by training our proposed network for one epoch with varying batch sizes. We chose one epoch as the main goal of this investigation is to find trends in the accuracy compared to the batch size. As can be seen in Table \ref{table:batch}, batch sizes of 256 and higher tend to have poor validation and testing accuracies. While a larger batch size can result in faster training time \cite{you2017imagenet}, this has to be balanced with reduced accuracy.

\textbf{Filter Size.} Similar to the conducted experiments on the impact of batch size, we also trained our proposed network for one epoch. This was done, as again we are looking for trends in the accuracy compared to filter sizes. As can be seen in Table \ref{table:filter}, when the number of filters is 4, we achieve the highest accuracy. This is also reflected in our final results, as 4 filters were used for the results shown in Sections \ref{sec:framebased} and \ref{sec:videoBased}. Increasing the number of filters to 8 results in significant decrease in accuracy ($\sim20\%$), however, once the number of filters is $\geq 256$, we see the accuracies converge to approximately $50\%$ on both training and validation. Because the larger batch sizes we tested ($\geq 256$) needed 1-3 days to train, we inferred the accuracy of the testing sets for these number of filters based on the accuracy of using 8 filters.

\begin{table}
  \centering
  \caption{Impact of number of filters on accuracy when training proposed network for 1 epoch. NOTE: * denotes inferred accuracy based on results from 8 filters, due to time required to train networks with larger filter sizes.}
    \setlength{\tabcolsep}{3pt}
    \begin{tabular}{|c|c|c|c|}
    \hline
    \makecell{Number of\\Filters} & \makecell{Training\\Accuracy} & \makecell{Validation\\ Accuracy} & \makecell{Testing\\Accuracy} \\
    \hline
    \bfseries 4 & \bfseries 0.7029 &  \bfseries 0.8080 & \bfseries 0.6919 \\
    \hline
    8 & 0.4999 &  0.5002 & 0.500 \\
    \hline
    256 & 0.5002 & 0.5002 & 0.500* \\
    \hline
    512 & 0.5002 & 0.5002 & 0.500* \\
    \hline
    1024 & 0.5002 & 0.5002 & 0.500* \\
    \hline
    \end{tabular}
    \label{table:filter}
\end{table}

\subsection{Comparisons to State of the Art}
R{\"o}ssler et al. \cite{rossler2018faceforensics} conducted experiments on the FaceForensics dataset using a subset of the available data. They selected 20 frames (10 original and 10 fake) from each of the $704$ available videos in the training set. They also selected 20 frames from each of the 150 validation and testing videos. To conduct their experiments, they used XceptionNet \cite{chollet2017xception} by freezing the first $36$ layers and replacing the last layer with a dense layer of 2 nodes (original and fake). They trained the network for 10 epochs using the Adam optimizer with a learning rate of 0.001 and a batch size of 64. They achieved an accuracy of $99.3\%$, compared to ours of $99.6\%$. 

As a subset of the data was used for this experiment, we are interested in how using XceptionNet, with these changes, would impact the accuracy on the entire training, validation, and testing sets as we did in our experimental design. Considering this, we replicated this experimental design using XceptionNet (freezing first 36 layers, new dense layer, training for 10 epochs) and trained on the entire training set. This resulted in an accuracy of $50\%$ for both the validation and testing sets. As can be seen in Table \ref{table:xception}, the training accuracy for this architecture was high ($> 99\%$) for all epochs, however, the validation and testing accuracies converged to $50\%$ for all epochs. These results further validate the robustness of using a smaller network, compared to a larger one, to detect fake face videos.
\begin{table}
  \centering
  \caption{Accuracies (per epoch) of replicated XceptionNet architecture on entire FaceForensics training, validation, and testing sets. NOTE: even numbered epochs shown for brevity, but all epochs had same validation/testing accuracies.}
    \setlength{\tabcolsep}{3pt}
    \begin{tabular}{|c|c|c|c|}
    \hline
    Epoch & \makecell{Training\\Accuracy} & \makecell{Validation\\ Accuracy} & \makecell{Testing\\Accuracy} \\
    \hline
    2 & 0.9987 &  0.5 & 0.5 \\
    \hline
    4 & 0.9993 &  0.5 & 0.5 \\
    \hline
    6 & 0.9995 & 0.5 & 0.5 \\
    \hline
    8 & 0.9996 & 0.5 & 0.5 \\
    \hline
    10 & 0.9997 & 0.5 & 0.5 \\
    \hline
    \end{tabular}
    \label{table:xception}
\end{table}
\section{Conclusion}
We proposed the use of a smaller (less parameters to learn) CNN for detecting fake face videos. We investigate both frame- and video-based approaches to this problem achieving accuracies of $99.6\%$ and $99.67\%$, respectively, on the FaceForensics dataset \cite{rossler2018faceforensics}. We show state-of-the-art results and validation of the chosen hyperparameters (e.g. batch size, number of filters) for the proposed network. Due to the increase, in recent years, of manipulated videos, this work has broader impacts in security and digital communication.
\section*{Acknowledgment}
This material is based on work that was supported in part by an Amazon Machine Learning Research Award.
\bibliographystyle{ieee}
\bibliography{references}

\begin{thebibliography}{10}\itemsep=-1pt

\bibitem{abu2016youtube}
S.~Abu-El-Haija, N.~Kothari, J.~Lee, P.~Natsev, G.~Toderici, B.~Varadarajan,
  and S.~Vijayanarasimhan.
\newblock Youtube-8m: A large-scale video classification benchmark.
\newblock {\em arXiv preprint arXiv:1609.08675}, 2016.

\bibitem{agarwal2019protecting}
S.~Agarwal, H.~Farid, Y.~Gu, M.~He, K.~Nagano, and H.~Li.
\newblock Protecting world leaders against deep fakes.
\newblock In {\em Proceedings of the IEEE Conference on Computer Vision and
  Pattern Recognition Workshops}, pages 38--45, 2019.

\bibitem{ba2014deep}
J.~Ba and R.~Caruana.
\newblock Do deep nets really need to be deep?
\newblock In {\em Advances in neural information processing systems}, pages
  2654--2662, 2014.

\bibitem{chollet2017xception}
F.~Chollet.
\newblock Xception: Deep learning with depthwise separable convolutions.
\newblock In {\em Proceedings of the IEEE conference on computer vision and
  pattern recognition}, pages 1251--1258, 2017.

\bibitem{FACS}
P.~Ekman and E.~Rosenberg.
\newblock What the face reveals: Basic and applied studies of spontaneous
  expression using the facial action coding system (facs).
\newblock {\em Oxford University Press}, 1997.

\bibitem{goodfellow2014generative}
I.~Goodfellow, J.~Pouget-Abadie, M.~Mirza, B.~Xu, D.~Warde-Farley, S.~Ozair,
  A.~Courville, and Y.~Bengio.
\newblock Generative adversarial nets.
\newblock In {\em Advances in neural information processing systems}, pages
  2672--2680, 2014.

\bibitem{guera2018deepfake}
D.~G{\"u}era and E.~J. Delp.
\newblock Deepfake video detection using recurrent neural networks.
\newblock In {\em 2018 15th IEEE International Conference on Advanced Video and
  Signal Based Surveillance (AVSS)}, pages 1--6. IEEE, 2018.

\bibitem{kingma2014adam}
D.~P. Kingma and J.~Ba.
\newblock Adam: A method for stochastic optimization.
\newblock {\em arXiv preprint arXiv:1412.6980}, 2014.

\bibitem{li2018ictu}
Y.~Li, M.-C. Chang, and S.~Lyu.
\newblock In ictu oculi: Exposing ai created fake videos by detecting eye
  blinking.
\newblock In {\em 2018 IEEE International Workshop on Information Forensics and
  Security (WIFS)}, pages 1--7. IEEE, 2018.

\bibitem{li2018exposing}
Y.~Li and S.~Lyu.
\newblock Exposing deepfake videos by detecting face warping artifacts.
\newblock {\em arXiv preprint arXiv:1811.00656}, 2018.

\bibitem{matern2019exploiting}
F.~Matern, C.~Riess, and M.~Stamminger.
\newblock Exploiting visual artifacts to expose deepfakes and face
  manipulations.
\newblock In {\em 2019 IEEE Winter Applications of Computer Vision Workshops
  (WACVW)}, pages 83--92. IEEE, 2019.

\bibitem{nguyen2019capsule}
H.~H. Nguyen, J.~Yamagishi, and I.~Echizen.
\newblock Capsule-forensics: Using capsule networks to detect forged images and
  videos.
\newblock In {\em ICASSP 2019-2019 IEEE International Conference on Acoustics,
  Speech and Signal Processing (ICASSP)}, pages 2307--2311. IEEE, 2019.

\bibitem{patel2015live}
K.~Patel, H.~Han, A.~K. Jain, and G.~Ott.
\newblock Live face video vs. spoof face video: Use of moir{\'e} patterns to
  detect replay video attacks.
\newblock In {\em 2015 International Conference on Biometrics (ICB)}, pages
  98--105. IEEE, 2015.

\bibitem{radiuk2017impact}
P.~M. Radiuk.
\newblock Impact of training set batch size on the performance of convolutional
  neural networks for diverse datasets.
\newblock {\em Information Technology and Management Science}, 20(1):20--24,
  2017.

\bibitem{rossler2018faceforensics}
A.~R{\"o}ssler, D.~Cozzolino, L.~Verdoliva, C.~Riess, J.~Thies, and
  M.~Nie{\ss}ner.
\newblock Faceforensics: A large-scale video dataset for forgery detection in
  human faces.
\newblock {\em arXiv preprint arXiv:1803.09179}, 2018.

\bibitem{sabour2017dynamic}
S.~Sabour, N.~Frosst, and G.~E. Hinton.
\newblock Dynamic routing between capsules.
\newblock In {\em Advances in neural information processing systems}, pages
  3856--3866, 2017.

\bibitem{suwajanakorn2017synthesizing}
S.~Suwajanakorn, S.~M. Seitz, and I.~Kemelmacher-Shlizerman.
\newblock Synthesizing obama: learning lip sync from audio.
\newblock {\em ACM Transactions on Graphics (TOG)}, 36(4):1--13, 2017.

\bibitem{thies2016face2face}
J.~Thies, M.~Zollhofer, M.~Stamminger, C.~Theobalt, and M.~Nie{\ss}ner.
\newblock Face2face: Real-time face capture and reenactment of rgb videos.
\newblock In {\em Proceedings of the IEEE conference on computer vision and
  pattern recognition}, pages 2387--2395, 2016.

\bibitem{yang2019exposing}
X.~Yang, Y.~Li, and S.~Lyu.
\newblock Exposing deep fakes using inconsistent head poses.
\newblock In {\em ICASSP 2019-2019 IEEE International Conference on Acoustics,
  Speech and Signal Processing (ICASSP)}, pages 8261--8265. IEEE, 2019.

\bibitem{you2017imagenet}
Y.~You, Z.~Zhang, J.~Demmel, K.~Keutzer, and C.-J. Hsieh.
\newblock Imagenet training in 24 minutes.
\newblock {\em arXiv preprint arXiv:1709.05011}, 2017.

\end{thebibliography}

\end{document}